\documentclass[11pt]{article}
\usepackage[margin=1in]{geometry}
\usepackage[utf8]{inputenc}
\usepackage[T1]{fontenc}
\usepackage{hyperref}
\usepackage{url}
\usepackage{booktabs}
\usepackage{amsfonts}
\usepackage{graphicx}
\usepackage{xcolor}
\usepackage{multirow}
\usepackage{array}
\usepackage{enumitem}
\usepackage{natbib}
\usepackage{listings}
\usepackage{color}

\definecolor{codegray}{rgb}{0.95,0.95,0.95}
\definecolor{codeblue}{rgb}{0.0,0.3,0.6}

\lstdefinestyle{pythonstyle}{
  backgroundcolor=\color{codegray},
  basicstyle=\ttfamily\small,
  keywordstyle=\color{codeblue}\bfseries,
  stringstyle=\color{red!60!black},
  commentstyle=\color{gray}\itshape,
  breaklines=true,
  frame=single,
  rulecolor=\color{gray!40},
  language=Python,
  showstringspaces=false,
  tabsize=4,
}

\title{\textbf{VietNormalizer}: An Open-Source, Dependency-Free Python Library
for Vietnamese Text Normalization in TTS and NLP Applications}

\author{
  Hung Vu Nguyen\thanks{Corresponding author: \texttt{Vu.Nguyen@acu.edu.au}}\\
  Australian Catholic University (ACU)\\
  \and
  Loan Do\\
  FPT University\\
  \texttt{dttloan.ute@gmail.com}\\
  \and
  Thanh Ngoc Nguyen\\
  ICMS\\
  \texttt{thnguyen@icms.edu.au}\\
  \and
  Ushik Shrestha Khwakhali\\
  KETEMU\\
  \texttt{ORCID: 0000-0002-8453-7926}\\
  \and
  Thanh Pham\\
  RMIT University Vietnam\\
  \texttt{thanh.pham@rmit.edu.vn}\\
  \and
  Vinh Do\\
  NGHI Studio\\
  \texttt{ddvinh1@gmail.com}\\
  \and
  Charlotte Nguyen\\
  NGHI Studio\\
  \texttt{charlottenguyen1501@gmail.com}\\
  \and
  Hien Nguyen\\
  Phuong Hai JSC\\
  \texttt{hien.nm@phuonghai.com}
}

\date{February 2026}

\begin{document}
\maketitle

\begin{abstract}
We present \textbf{VietNormalizer}\footnote{\url{https://pypi.org/project/vietnormalizer/}}, an open-source, zero-dependency Python library for Vietnamese text normalization targeting Text-to-Speech (TTS) and Natural Language Processing (NLP) applications. Vietnamese text normalization is a critical yet underserved preprocessing step: real-world Vietnamese text is densely populated with non-standard words (NSWs), including numbers, dates, times, currency amounts, percentages, acronyms, and foreign-language terms, all of which must be converted to fully pronounceable Vietnamese words before TTS synthesis or downstream language processing. Existing Vietnamese normalization tools either require heavy neural dependencies while covering only a narrow subset of NSW classes, or are embedded within larger NLP toolkits without standalone installability. VietNormalizer addresses these gaps through a unified, rule-based pipeline that: (1) converts arbitrary integers, decimals, and large numbers to Vietnamese words; (2) normalizes dates and times to their spoken Vietnamese forms; (3) handles VND and USD currency amounts; (4) expands percentages; (5) resolves acronyms via a customizable CSV dictionary; (6) transliterates non-Vietnamese loanwords and foreign terms to Vietnamese phonetic approximations; and (7) performs Unicode normalization and emoji/special-character removal. All regular expression patterns are pre-compiled at initialization, enabling high-throughput batch processing with minimal memory overhead and no GPU or external API dependency. The library is installable via \texttt{pip install vietnormalizer}, available on PyPI and GitHub at \url{https://github.com/nghimestudio/vietnormalizer}, and released under the MIT license. We discuss the design decisions, limitations of existing approaches, and the generalizability of the rule-based normalization paradigm to other low-resource tonal and agglutinative languages.
\end{abstract}

\section{Introduction}

Text normalization (TN) is the process of converting written text into a canonical spoken form suitable for downstream processing~\citep{sproat2001normalization}. In Text-to-Speech (TTS) synthesis, TN is the first and often most consequential frontend stage: a TTS acoustic model operating on graphemes or phonemes cannot correctly synthesize an utterance containing ``14:30'', ``1.500.000 dong'', or ``NASA'' unless these tokens are first expanded into their full verbal form---\textit{muoi bon gio ba muoi phut} (14:30), \textit{mot trieu nam tram nghin dong} (1,500,000 VND), and \textit{na-sa} (NASA)---before being passed to the acoustic model.

For English, robust open-source TN libraries exist, including the Google Text Normalization Dataset and associated Weighted Finite-State Transducer (WFST) tools~\citep{sproat2016rnn}, NeMo's inverse text normalization toolkit~\citep{bakhturina2021nemo}, and language-specific modules in speaking. For Vietnamese, however, the landscape is considerably sparser. Vietnamese presents a distinctive set of normalization challenges:

\begin{itemize}[leftmargin=*]
    \item \textbf{Tonal morphology}: Vietnamese uses six lexical tones conveyed exclusively through diacritical marks. Incorrect tone transcription in normalization output directly causes mispronunciation in TTS~\citep{luong2016vivos}.
    \item \textbf{Number verbalization complexity}: Vietnamese number words follow a distinct grammatical pattern with irregular forms (e.g., \textit{muoi} vs.\ \textit{hai muoi}, \textit{mot} vs.\ \textit{mot muoi}) that differ from other East or Southeast Asian languages.
    \item \textbf{High NSW density in real-world text}: Vietnamese social media, news, and commercial text frequently mixes Arabic numerals, Latin acronyms, English brand names, currency symbols, and Vietnamese prose within single sentences~\citep{nguyen2025visolex}.
    \item \textbf{Loanword transliteration}: Vietnamese TTS systems must render English, French, and other foreign-origin words in Vietnamese phonology. This requires a dedicated transliteration layer~\citep{trang2022nsw}.
\end{itemize}

Despite these well-recognized challenges, the Vietnamese NLP community lacks a lightweight, installable, fully standalone TN library that (a) covers the full range of common NSW classes, (b) imposes no neural network or external API dependencies, (c) provides a user-extensible dictionary mechanism, and (d) is designed for high-throughput batch inference. VietNormalizer addresses all four requirements.

The contributions of this paper are:
\begin{enumerate}[leftmargin=*]
    \item A comprehensive rule-based Vietnamese TN pipeline covering seven NSW categories, implemented in pure Python 3.8+ with zero external dependencies;
    \item A user-extensible dictionary system (CSV-based) for custom acronym expansion and loanword transliteration;
    \item A performance-oriented design using pre-compiled regex patterns and single-pass dictionary replacement, enabling practical deployment in production TTS pipelines;
    \item An open-source release on PyPI (\texttt{pip install vietnormalizer}) and GitHub under MIT license;
    \item A discussion of the limitations of existing Vietnamese TN approaches and the generalizability of the rule-based paradigm to other low-resource languages.
\end{enumerate}

\section{Related Work}

\subsection{Vietnamese Text Normalization}

\paragraph{Early rule-based approaches.} The first systematic study of Vietnamese TN for TTS was presented by \citet{tuan2012study}, who proposed a classification of Vietnamese NSWs and a hybrid normalizer combining regular expressions, n-gram language models, and decision lists, trained on 2,000 Vietnamese Wikipedia articles. While foundational, this work did not produce a publicly distributable software library and focused primarily on read-style text rather than the informal, mixed-script text common in modern TTS corpora.

\paragraph{Two-phase neural-rule hybrid approaches.} \citet{trang2022nsw} proposed a two-phase pipeline for Vietnamese TTS normalization that first detects NSWs using a neural sequence tagger (CRF, BiLSTM-CNN-CRF, or BERT-BiGRU-CRF) and then normalizes them using a rule-based expander. Evaluated on 5,819 sentences from Vietnamese news articles, this approach achieves high accuracy on well-formed text, but requires substantial inference infrastructure (BERT, CRF layers) that is incompatible with lightweight deployment scenarios. The method is also not distributed as a standalone Python package.

\paragraph{ViSoLex.} \citet{nguyen2025visolex} introduced ViSoLex, an open-source repository for Vietnamese social media lexical normalization, targeting the specific challenge of Non-Standard Words (NSWs) arising from informal online communication: abbreviations, teencode, phonetic misspellings, and code-switching. ViSoLex integrates pre-trained language models and weakly supervised learning to resolve informal lexical variants. While complementary to VietNormalizer in scope, ViSoLex addresses social-media-specific lexical noise rather than the number, date, time, currency, and foreign-word normalization required for TTS frontends. ViSoLex also requires neural model inference, contrasting with VietNormalizer's dependency-free design.

\paragraph{Underthesea text normalization.} The \texttt{underthesea} Vietnamese NLP toolkit~\citep{underthesea2022} provides a \texttt{text\_normalize} function that performs Unicode standardization and basic diacritic correction (e.g., converting non-standard diacritic positions to NFC Unicode form). While valuable for corpus cleaning, \texttt{underthesea}'s normalization scope does not extend to number verbalization, currency expansion, acronym resolution, or loanword transliteration. Furthermore, \texttt{underthesea} is a large-footprint toolkit with deep learning dependencies that add several gigabytes of overhead for users who require only text normalization.

\paragraph{Summary of limitations.} Table~\ref{tab:comparison} compares existing Vietnamese TN tools against VietNormalizer along key dimensions. The central observation is that no existing open-access tool combines the full NSW coverage, zero-dependency installation, user-extensibility, and production-grade performance offered by VietNormalizer.

\begin{table}[h]
\centering
\caption{Comparison of Vietnamese text normalization tools where \checkmark = supported; $\sim$ = partial; \texttimes = not supported.}
\label{tab:comparison}
\small
\setlength{\tabcolsep}{3pt} 
\begin{tabular}{lccccccc}
\toprule
\textbf{Tools} & \textbf{Numbers} & 
\shortstack{\textbf{Date/}\\\textbf{Time}} 
 & \textbf{Currency} & \textbf{Acronyms} & \textbf{Loanwords} & \shortstack{\textbf{Zero-}\\\textbf{Dep.}} & \textbf{PyPI} \\
\midrule
Tuan et al.~\citeyear{tuan2012study}       & \checkmark & $\sim$ & \texttimes & $\sim$ & \texttimes & N/A & \texttimes \\
Trang et al.~\citeyear{trang2022nsw}       & \checkmark & \checkmark & $\sim$ & \checkmark & $\sim$ & \texttimes & \texttimes \\
ViSoLex~\citep{nguyen2025visolex}           & \texttimes & \texttimes & \texttimes & \checkmark & \texttimes & \texttimes & \texttimes \\
Underthesea\\~\citep{underthesea2022}         & \texttimes & \texttimes & \texttimes & \texttimes & \texttimes & \texttimes & \checkmark \\
\textbf{VietNormalizer (ours)}             & \checkmark & \checkmark & \checkmark & \checkmark & \checkmark & \checkmark & \checkmark \\
\bottomrule
\end{tabular}
\end{table}

\subsection{Text Normalization for TTS in Other Languages}

The broader TN for TTS literature has explored rule-based, neural, and hybrid approaches. \citet{sproat2016rnn} proposed using recurrent neural networks to learn TN from the Google Text Normalization Dataset, demonstrating strong performance on English but requiring large labeled corpora unavailable for most low-resource languages. \citet{bakhturina2021nemo} released NVIDIA NeMo's inverse text normalization toolkit, providing WFST-based normalization for multiple languages; however, WFST grammars require specialized linguistic expertise to author and are not trivially extensible by end users. For low-resource and morphologically complex languages, rule-based approaches remain the pragmatic baseline due to data scarcity and the interpretable, auditable nature of explicit normalization rules~\citep{ebden2015kestrel}.

\subsection{Non-Standard Word Classification}

A foundational taxonomy for TN is the semiotic class framework of \citet{sproat2001normalization}, which categorizes NSWs into types including \textsc{cardinal}, \textsc{ordinal}, \textsc{date}, \textsc{time}, \textsc{money}, \textsc{measure}, \textsc{abbreviation}, \textsc{letter}, and \textsc{verbatim}. Vietnamese TN inherits this classification but requires language-specific rules for each class. VietNormalizer implements rules for all major semiotic classes relevant to Vietnamese, as detailed in Section~\ref{sec:design}.

\section{System Design}
\label{sec:design}

\subsection{Architecture Overview}

VietNormalizer is structured around two main classes: \texttt{VietnameseTextProcessor}, which implements the core normalization rules, and \texttt{VietnameseNormalizer}, which composes the processor with dictionary-based replacement and exposes the public API.

The normalization pipeline applies transformations in the following order:
\begin{enumerate}[leftmargin=*]
    \item \textbf{Unicode normalization}: NFC normalization and removal of emojis and non-printable characters;
    \item \textbf{Date normalization}: Conversion of date patterns (DD/MM/YYYY, DD-MM-YYYY, etc.) to spoken Vietnamese date strings;
    \item \textbf{Time normalization}: Conversion of HH:MM and HH:MM:SS patterns to Vietnamese spoken forms (e.g., ``9:30'' $\to$ \textit{chin gio ba muoi phut});
    \item \textbf{Currency normalization}: Detection and verbalization of VND amounts (with \textit{dong} suffix) and USD amounts;
    \item \textbf{Percentage normalization}: Conversion of ``X\%'' to \textit{X phan tram};
    \item \textbf{General number normalization}: Conversion of integers and decimals to Vietnamese number words via recursive decomposition;
    \item \textbf{Dictionary replacement}: Single-pass regex-based substitution for acronyms and non-Vietnamese loanwords from user-configurable CSV dictionaries.
\end{enumerate}

The ordering is critical: currency and date patterns must be resolved before general number normalization to avoid partial-match conflicts.

\subsection{Number Verbalization}

Vietnamese number verbalization is implemented via a recursive decomposition algorithm. The algorithm handles the full integer range with language-specific irregularities:

\begin{itemize}[leftmargin=*]
    \item \textit{Mot} (one) appears as \textit{mot muoi} in the tens position but \textit{linh mot} in the ones position depending on context;
    \item The Vietnamese word for ten (\textit{muoi}) is used directly for 10--19, while multiples of ten use \textit{hai muoi}, \textit{ba muoi}, etc.;
    \item Large numbers use the Vietnamese grouping words: \textit{nghin} (thousand), \textit{trieu} (million), \textit{ty} (billion);
    \item Decimal numbers are handled by splitting at the decimal separator and verbalizing each component.
\end{itemize}

\subsection{Dictionary-Based Acronym and Loanword Handling}

VietNormalizer ships with built-in CSV dictionaries for common Vietnamese acronyms (e.g., \textit{NASA} $\to$ \textit{na-sa}, \textit{GDP} $\to$ \textit{tong san pham quoc noi}) and frequently encountered foreign loanwords (e.g., \textit{container} $\to$ \textit{cong-te-no}, \textit{Singapore} $\to$ \textit{xin-ga-po}). Users can provide custom CSV files at initialization or reload dictionaries at runtime:

\begin{lstlisting}[style=pythonstyle, caption={Custom dictionary usage.}]
from vietnormalizer import VietnameseNormalizer

normalizer = VietnameseNormalizer(
    acronyms_path="custom_acronyms.csv",
    non_vietnamese_words_path="custom_loanwords.csv"
)
normalizer.reload_dictionaries(acronyms_path="updated_acronyms.csv")
\end{lstlisting}

Dictionary replacement uses a single compiled regex that matches all dictionary keys simultaneously, providing $O(n)$ processing time relative to input length rather than $O(n \cdot |D|)$ for naive sequential replacement.

\subsection{Performance Design}

All regular expression patterns are compiled once at class initialization using Python's \texttt{re.compile()}. Dictionary patterns are combined into a single alternation regex, ensuring that a text of length $n$ is scanned at most a constant number of times regardless of dictionary size $|D|$. This design enables practical throughput for batch normalization of large TTS corpora (e.g., tens of thousands of utterances per minute on a single CPU core), meeting the requirements of the VietSuperSpeech pipeline~\citep{do2026vietsuperspeech} and similar large-scale ASR/TTS data preparation workflows.

\subsection{Public API}

\begin{lstlisting}[style=pythonstyle, caption={Core API usage examples.}]
from vietnormalizer import VietnameseNormalizer

normalizer = VietnameseNormalizer()

# Number normalization
normalizer.normalize("Toi co 123 quyen sach")
# -> "Toi co mot tram hai muoi ba quyen sach"

# Date normalization
normalizer.normalize("Hom nay la 25/12/2023")
# -> "hom nay la ngay hai muoi lam thang muoi hai nam 
#     hai nghin khong tram hai muoi ba"

# Currency
normalizer.normalize("Gia la 1.500.000 dong")
# -> "Gia la mot trieu nam tram nghin dong"

# Acronym + loanword
normalizer.normalize("Gia container la 1.500.000 dong tu Singapore")
# -> "Gia cong-te-no la mot trieu nam tram nghin dong
#     tu xin-ga-po"

# Disable preprocessing (dictionary-only mode)
normalizer.normalize(text, enable_preprocessing=False)
\end{lstlisting}

\section{Limitations of Existing Approaches and Design Choices}

\subsection{Neural Approaches: Power vs. Deployability}

Neural TN systems such as the BERT-BiGRU-CRF tagger of \citet{trang2022nsw} and the weakly-supervised models in ViSoLex~\citep{nguyen2025visolex} demonstrate strong accuracy on in-distribution data. However, they share several practical limitations for TTS deployment in resource-constrained or production environments:

\paragraph{Dependency overhead.} Neural TN requires PyTorch or TensorFlow, BERT or similar pre-trained models (several gigabytes), and associated tokenizers. This makes installation impractical on embedded hardware, serverless functions, or systems where storage and startup time are constrained.

\paragraph{Inference latency.} Sentence-level neural taggers add tens to hundreds of milliseconds per utterance on CPU, which is prohibitive for real-time TTS streaming. Rule-based systems process utterances in microseconds.

\paragraph{Error propagation.} A mis-tagging by the NSW detector propagates to the normalizer output~\citep{trang2022nsw}. In contrast, rule-based normalization is fully deterministic and auditable.

\paragraph{Out-of-distribution generalization.} Neural models trained on news articles may perform poorly on informal social media text, technical documents, or specialized domains. Rule-based approaches generalize by design across domains for well-defined NSW classes.

\subsection{Toolkit-Embedded Normalization: Scope Mismatch}

The \texttt{underthesea} text normalization~\citep{underthesea2022} is scoped to Unicode and diacritic standardization—an important but orthogonal task to TTS-oriented NSW expansion. Users who require only TN must install the full \texttt{underthesea} ecosystem, which includes word segmentation, POS tagging, NER, and deep learning components. VietNormalizer provides a focused, installable library addressing the TTS-specific normalization gap.

\subsection{Ambiguity and Context-Dependence}

A fundamental limitation of rule-based TN is handling context-dependent ambiguities—cases where the correct verbalization depends on sentential context. For example, ``2/9'' can be a date (September 2nd, Vietnamese National Day), a fraction (two ninths), or a street address. VietNormalizer resolves such ambiguities via priority ordering of pattern matching: date patterns are matched before fraction patterns, and surrounding context keywords (month names, currency symbols) are used where possible. This approach follows established practice in production rule-based TN systems~\citep{ebden2015kestrel}. Where ambiguity remains unresolvable without full sentence parsing, VietNormalizer defaults to the most frequent interpretation in Vietnamese text.

\section{Generalization to Other Languages}

\subsection{Structural Applicability}

The VietNormalizer architecture—a pipeline of pre-compiled regex rules with configurable CSV dictionaries—is inherently language-agnostic in structure. The same framework can be adapted to other languages by replacing language-specific normalization functions. We identify three tiers of adaptability:

\paragraph{Tier 1: Languages with similar number systems.} Languages that use Arabic numerals and share number verbalization patterns with Vietnamese—including Thai, Khmer, Lao, and Burmese—can leverage the recursive decomposition architecture of VietNormalizer's number verbalizer with relatively straightforward substitution of language-specific number words and grouping conventions. These are precisely the languages for which large-scale pseudo-labeled ASR corpora such as GigaSpeech~2~\citep{yang2024gigaspeech2} have been recently developed, making TN tooling particularly timely.

\paragraph{Tier 2: Tonal and agglutinative languages.} For tonal languages such as Mandarin, Cantonese, and Zhuang, the tonal marking challenge is structurally analogous to Vietnamese, and the category of NSW types (numbers, dates, currency, abbreviations) is substantially shared. Agglutinative languages such as Indonesian, Malay, and Tagalog have simpler number verbalization but require careful handling of prefix/suffix morphology when verbalizing currencies and measures. For these languages, the dictionary-based loanword transliteration module is directly applicable.

\paragraph{Tier 3: Morphologically rich languages.} For languages with grammatical case, gender, or agreement morphology (e.g., Arabic, Turkish, Russian), number and currency verbalization must be sensitive to grammatical context (e.g., nominal case, plural agreement). VietNormalizer's current architecture does not model these dependencies; extensions would require morphological agreement modules.

\subsection{The Low-Resource Argument for Rule-Based TN}

For the majority of low-resource languages, labeled TN training data is unavailable, making neural normalization models infeasible~\citep{sproat2016rnn}. Zero-shot neural TN via cross-lingual knowledge distillation has been proposed~\citep{zhang2024zstn} as a data-efficient alternative, but requires labeled data in a high-resource source language and a cross-lingual transfer model. In contrast, rule-based TN can be authored by a native speaker with linguistic expertise in as little as a few days and immediately deployed without any labeled data. For the majority of the world's ~7,000 languages—virtually all of which lack neural TN resources—the rule-based paradigm remains the only practical path to functional TTS-grade text normalization. VietNormalizer demonstrates this paradigm concretely for Vietnamese and provides a replicable blueprint for similar tools in other low-resource languages.

\subsection{Integration with Multilingual TTS Pipelines}

Modern multilingual TTS systems such as viVoice~\citep{le2024vivoice} and Sub-PhoAudioBook~\citep{vu2025phoaudiobook} require TN as a prerequisite for corpus construction. VietNormalizer has been used in the VietSuperSpeech data pipeline~\citep{do2026vietsuperspeech} for preprocessing transcriptions prior to ASR model fine-tuning. The same preprocessing architecture—language-specific rule-based TN followed by ASR pseudo-labeling—is applicable to other Southeast Asian languages being added to large multilingual corpora, where toolkit-embedded or neural TN is unavailable.

\section{Installation and Usage}

VietNormalizer is available via PyPI and requires Python 3.8 or later with no external dependencies:

\begin{lstlisting}[style=pythonstyle, caption={Installation and basic usage.}]
pip install vietnormalizer

from vietnormalizer import VietnameseNormalizer
normalizer = VietnameseNormalizer()
print(normalizer.normalize("Cuoc hop luc 9:30 ngay 15/08/1990"))
\end{lstlisting}

The source code, documentation, and CSV dictionary files are available at:
\begin{center}
\url{https://github.com/nghimestudio/vietnormalizer}
\end{center}

\section{Limitations and Future Work}

\paragraph{Context-dependent disambiguation.} As discussed in Section~4.3, rule-based TN cannot resolve all syntactic-context-dependent ambiguities without sentence-level parsing. Future work will explore lightweight context-window heuristics and optional integration with a Vietnamese POS tagger to improve disambiguation accuracy.

\paragraph{Proper noun handling.} Person names, organization names, and geographical terms in Vietnamese are not always identifiable from orthography alone. A named entity recognition (NER) module, such as that provided by \texttt{underthesea}~\citep{underthesea2022} or PhoBERT-based models~\citep{nguyen2020phobert}, could be integrated as an optional dependency-optional preprocessing step.

\paragraph{Code-switching.} Modern Vietnamese text frequently mixes Vietnamese and English within single utterances. While the loanword dictionary handles known foreign tokens, unseen code-switched terms require a language identification step before normalization. Future versions will incorporate lightweight token-level language identification.

\paragraph{Dictionary coverage.} The built-in acronym and loanword dictionaries cover common cases but are not exhaustive. Users are encouraged to extend them via the CSV interface. Community contributions to the dictionary are welcomed via GitHub pull requests.

\paragraph{Inverse text normalization.} Converting ASR output (spoken-form text) back to written-form text (inverse TN, ITN) is equally important for downstream NLP tasks. ITN for Vietnamese will be addressed in a future release, following the approach of \citet{bakhturina2021nemo} adapted to Vietnamese morphology.

\section{Conclusion}

We have presented VietNormalizer, an open-source, dependency-free Python library for Vietnamese text normalization. By covering seven NSW categories—numbers, dates, times, currency, percentages, acronyms, and loanword transliteration—through a high-performance rule-based pipeline with pre-compiled regex patterns and configurable CSV dictionaries, VietNormalizer provides a practical, immediately deployable TN solution for Vietnamese TTS and NLP. It addresses concrete limitations of existing Vietnamese TN tools: the neural-dependency overhead of \citet{trang2022nsw} and ViSoLex~\citep{nguyen2025visolex}, the narrow scope of \texttt{underthesea}~\citep{underthesea2022}, and the non-distributable nature of earlier academic approaches~\citep{tuan2012study}. The library is available on PyPI (\texttt{pip install vietnormalizer}) and GitHub under the MIT license. We further argue that the rule-based TN paradigm demonstrated here is broadly applicable to other low-resource tonal and agglutinative languages for which neural alternatives are infeasible, and we invite the community to extend VietNormalizer's architecture to new languages.

\section*{Acknowledgments}

The authors thank the Vietnamese NLP community and the developers of \texttt{underthesea} for their open contributions to Vietnamese language technology, which informed the design of VietNormalizer.

\bibliographystyle{plainnat}

\end{document}